\documentclass[runningheads]{llncs}
\usepackage[T1]{fontenc}
\usepackage{hyperref}
\usepackage{url}

\usepackage{graphicx}
\usepackage{multirow}
\usepackage{adjustbox}
\usepackage{bm}
\usepackage{makecell}
\usepackage{booktabs}
\usepackage{amsmath}
\begin{document}
%
\title{UAVDB: Point-Guided Masks for UAV Detection and Segmentation}
%
%
\author{Yu-Hsi Chen\orcidID{0009-0006-1771-0289}}
\authorrunning{Y.-H. Chen}
%
\institute{The University of Melbourne, Parkville, Australia\\
\email{yuhsi@student.unimelb.edu.au}\\
\url{https://github.com/wish44165/UAVDB}}
\maketitle              
\begin{abstract}
Accurate detection of Unmanned Aerial Vehicles (UAVs) is critical for surveillance, security, and airspace monitoring. However, existing datasets remain limited in scale, resolution, and the ability to capture objects across extreme size variations. To address these challenges, we present UAVDB, a benchmark dataset for UAV detection and segmentation, constructed via a point-guided weak supervision pipeline. We introduce Patch Intensity Convergence (PIC), a lightweight annotation method that converts trajectory points into bounding boxes, eliminating the need for manual labeling while preserving precise spatial localization. Building upon these annotations, we further generate segmentation masks using SAM2, enriching the dataset with multi-task labels. UAVDB consists of RGB frames from a fixed-camera multi-view video dataset, capturing UAVs across scales ranging from clearly visible objects to near single-pixel instances under diverse conditions. Quantitative results show that PIC combined with SAM2 outperforms existing annotation techniques in terms of IoU. Furthermore, we benchmark YOLO-based detectors on UAVDB, establishing baselines for future research.

\end{abstract}
%
%
%
\section{Introduction}
\label{sec:intro}
Precise UAV detection is crucial for effective monitoring and timely response to aerial threats. Despite advances in object detection, the performance of modern detectors remains highly dependent on the availability of high-quality annotations. In practice, this dependency becomes a major bottleneck, as noisy labels and missing instances substantially degrade performance, especially for tiny or rapidly moving UAVs.
Existing UAV-related datasets generally fall into two broad categories. The first focuses on ground-target detection, where aerial imagery is used to detect objects such as vehicles or pedestrians~\cite{ding2021object,wang2021tiny,xu2022detecting,zhu2021detection}. The second category comprises UAV-target datasets, where the UAV itself is the object of interest for detection or tracking. UAV-target datasets can be further divided into two subtypes:
(1) UAV-to-UAV datasets, in which a camera mounted on one UAV tracks another in flight~\cite{guo2025yolomg,li2016multi,Airborne_Object_Tracking_AWS,rozantsev2015flying}. These datasets require lots of operational effort, as they involve flying multiple UAVs simultaneously and precisely locating target UAVs, making the data collection process time-consuming and skill-intensive. (2) Camera-to-UAV datasets, where the UAV is observed by an external camera that may be handheld, mobile, or fixed (but not on a UAV), including both RGB~\cite{aksoy2019drone,kashiyama2020sky,pawelczyk2020real,steininger2021aircraft} and infrared~\cite{TGRS23OSCAR,huang2023anti,jiang2021anti,zhu2023evidential} modalities.

\begin{figure}[tb]
    \begin{center}
    \includegraphics[width=0.7\linewidth]{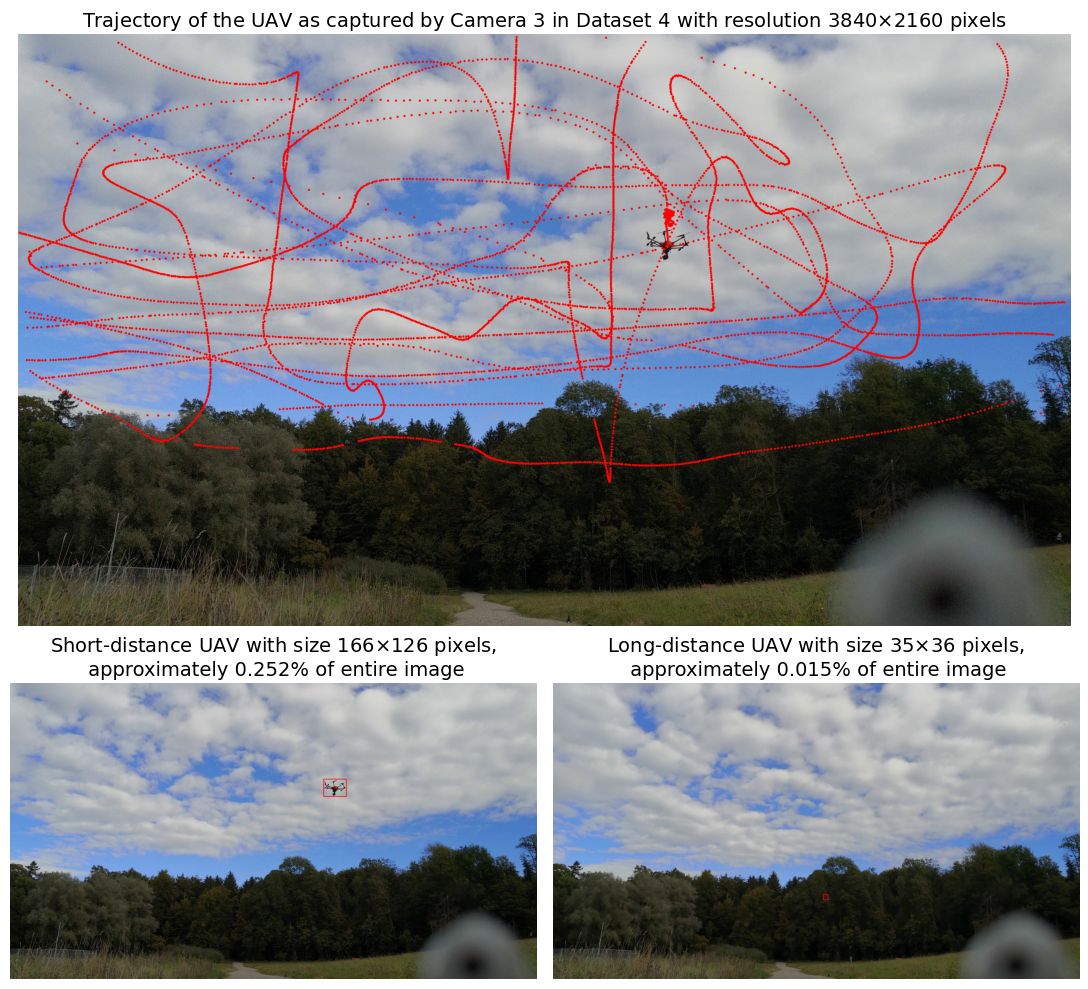}
    \end{center}
    \caption{UAV trajectory captured by Camera 3 in Dataset 4 at 3840\(\times\)2160 resolution in~\cite{li2020reconstruction}. The red path represents the UAV's trajectory. On the left, the UAV appears at a short distance with a size of 166\(\times\)126 pixels, occupying approximately 0.252\% of the total image area. On the right, the UAV is shown at a long distance, with a size of 35\(\times\)36 pixels, covering approximately 0.015\% of the entire image. This figure shows the varying visibility of the UAV depending on its distance from the camera.}
    \label{fig:trajectory}
\end{figure}

While several RGB-based camera-to-UAV datasets have been introduced in recent years~\cite{mta-rwowu_dataset,drone-9ab2n_dataset,mobile-net_dataset,drone_dataset}, they exhibit key limitations that hinder their applicability to real-world aerial surveillance, particularly for detecting small and distant UAVs in complex environments. These shortcomings underscore the need for a more representative and scalable benchmark.
For example, the dataset in~\cite{kashiyama2020sky} contains 600$\times$600 images but suffers from severe class imbalance (e.g., 74 birds vs. 1,392 helicopters), leading to biased learning, and is derived from low-frame-rate sequences unsuitable for temporal modeling. The dataset in~\cite{pawelczyk2020real} includes high-resolution videos (up to 4K), yet all data are downscaled to 640$\times$480, limiting the detection of tiny UAVs. Another dataset~\cite{steininger2021aircraft} spans resolutions from 192$\times$144 to 3840$\times$2160, but partial data inaccessibility undermines reproducibility. Additionally, datasets such as~\cite{aksoy2019drone,drone_dataset} lack temporal continuity due to non-sequential image collection, restricting motion-based analysis.
Overall, existing datasets often lack high-resolution temporal data, diverse environmental conditions, and consistent annotation quality. They predominantly feature large UAVs captured from short-range or ground-level viewpoints, which differ from real-world surveillance scenarios where UAVs are typically small, distant, and partially occluded in cluttered scenes.

\begin{table*}[tb]
\caption{Summary of dataset characteristics in~\cite{li2020reconstruction}. The table displays the number of frames and resolution for each camera across different datasets. Each cell lists the number of frames followed by the resolution in pixels.}
\label{tab:dataset}
\begin{center}
\begin{adjustbox}{max width=\linewidth}
\setlength{\tabcolsep}{8pt}
\begin{tabular}{cccccc}
\toprule
 Camera $\backslash$ Dataset & 1 & 2 & 3 & 4 & 5 \\
\midrule
 0 & 5334 / 1920$\times$1080 & 4377 / 1920$\times$1080 & 33875 / 1920$\times$1080 & 31075 / 1920$\times$1080 & 20970 / 1920$\times$1080 \\
 1 & 4941 / 1920$\times$1080 & 4749 / 1920$\times$1080 & 19960 / 1920$\times$1080 & 15409 / 1920$\times$1080 & 28047 / 1920$\times$1080 \\
 2 & 8016 / 1920$\times$1080 & 8688 / 1920$\times$1080 & 17166 / 3840$\times$2160& 15678 / 1920$\times$1080 & 31860 / 2704$\times$2028 \\
 3 & 4080 / 1920$\times$1080 & 4332 / 1920$\times$1080 & 14196 / 1440$\times$1080& 10933 / 3840$\times$2160 & 31992 / 1920$\times$1080 \\
 4 & -- & -- & 18900 / 1920$\times$1080 & 17640 / 1920$\times$1080 & 21523 / 2288$\times$1080\\
 5 & -- & -- & 28080 / 1920$\times$1080 & 32016 / 1920$\times$1080 & 17550 / 1920$\times$1080 \\
 6 & -- & -- & -- & 11292 / 1440$\times$1080 & --\\
\bottomrule
\end{tabular}
\end{adjustbox}
\end{center} 
\end{table*}

To overcome the limitations of existing RGB-based camera-to-UAV datasets, we introduce UAVDB, a high-resolution benchmark for multiscale UAV detection under diverse and challenging conditions captured by static ground-based cameras. Designed for long-range aerial surveillance, UAVDB focuses on detecting small and distant targets in realistic scenarios, such as monitoring restricted zones and critical infrastructure, thereby enabling evaluation under real-world operational constraints.
UAVDB is built upon the multi-view drone tracking dataset~\cite{li2020reconstruction}, originally developed for 3D trajectory reconstruction using unsynchronized consumer cameras with unknown viewpoints. It provides high-resolution RGB videos with corresponding 2D UAV locations, forming a strong foundation for annotation generation.
We propose Patch Intensity Convergence (PIC), which converts trajectory points into 2D bounding boxes, thereby eliminating the need for manual labeling. These boxes are then used as prompts for the SAM2~\cite{ravi2024sam} to generate instance masks, forming a fully automated pipeline from trajectory points to segmentation annotations.
To illustrate dataset diversity, we visualize representative UAV trajectories and human-labeled bounding boxes across different scale ranges, as shown in Fig.~\ref{fig:trajectory}, and summarize the multi-view drone tracking dataset~\cite{li2020reconstruction} details in Tab.~\ref{tab:dataset}.
Our contributions are summarized as follows:

\begin{enumerate}
    \item[$1.$] We introduce UAVDB, a high-resolution RGB video dataset for UAV detection and segmentation, featuring multiscale targets in complex and diverse environments. UAVDB is constructed by transforming trajectory data~\cite{li2020reconstruction} into precise bounding box annotations via the proposed PIC method, followed by applying SAM2~\cite{ravi2024sam} to generate instance masks.
    \item[$2.$] We validate the effectiveness and efficiency of the proposed PIC method in terms of IoU accuracy and runtime performance. In addition, we provide a comprehensive benchmark on UAVDB using YOLO-series detectors, ranging from YOLOv8 to YOLO26~\cite{Jocher_YOLO_by_Ultralytics_2023,wang2024yolov9,wang2024yolov10,yolo11_ultralytics,tian2025yolov12attentioncentricrealtimeobject,lei2025yolov13,yolo26_ultralytics}.
\end{enumerate}
%
%
%
\section{Related Work}
\label{sec:related_work}

\subsection{Point-Guided Weak Supervision}
\label{ssec:points}
Recent studies have demonstrated the effectiveness of point-level annotations as a weak form of supervision across various vision tasks. In object detection, including oriented object detection, single-point supervision has been explored as an alternative to bounding box annotations~\cite{li2023monte,luo2024pointobb,zhang2022group,ge2023point,chen2021points,ying2023mapping}, reducing annotation cost but often relying on complex pipelines such as point-to-box regression or handcrafted geometric priors. 
In instance segmentation, point annotations have been used to supervise mask generation~\cite{chen2025p2object,kim2023devil}, refine boundaries~\cite{breznik2024leveraging}, or guide proposal generation~\cite{yao2024position}; however, performance often degrades for small or irregular objects without additional supervision. In 3D object detection, point priors have also been leveraged to bridge 2D observations and 3D reasoning~\cite{gao2024leveraging}, but require multimodal inputs and task-specific architectural design.
Despite these advances, existing methods often rely on end-to-end training, exhibit limited cross-domain generalization, or incur high computational complexity. In contrast, we propose a training-free and plug-and-play pipeline that directly operates on trajectory points and raw video frames, enabling scalable and robust annotation generation without model retraining or domain-specific tuning.

\subsection{Bounding Box Extraction via Segmentation}
\label{ssec:segmentation}
\begin{figure*}[tb]
    \begin{center}
    \includegraphics[width=\linewidth]{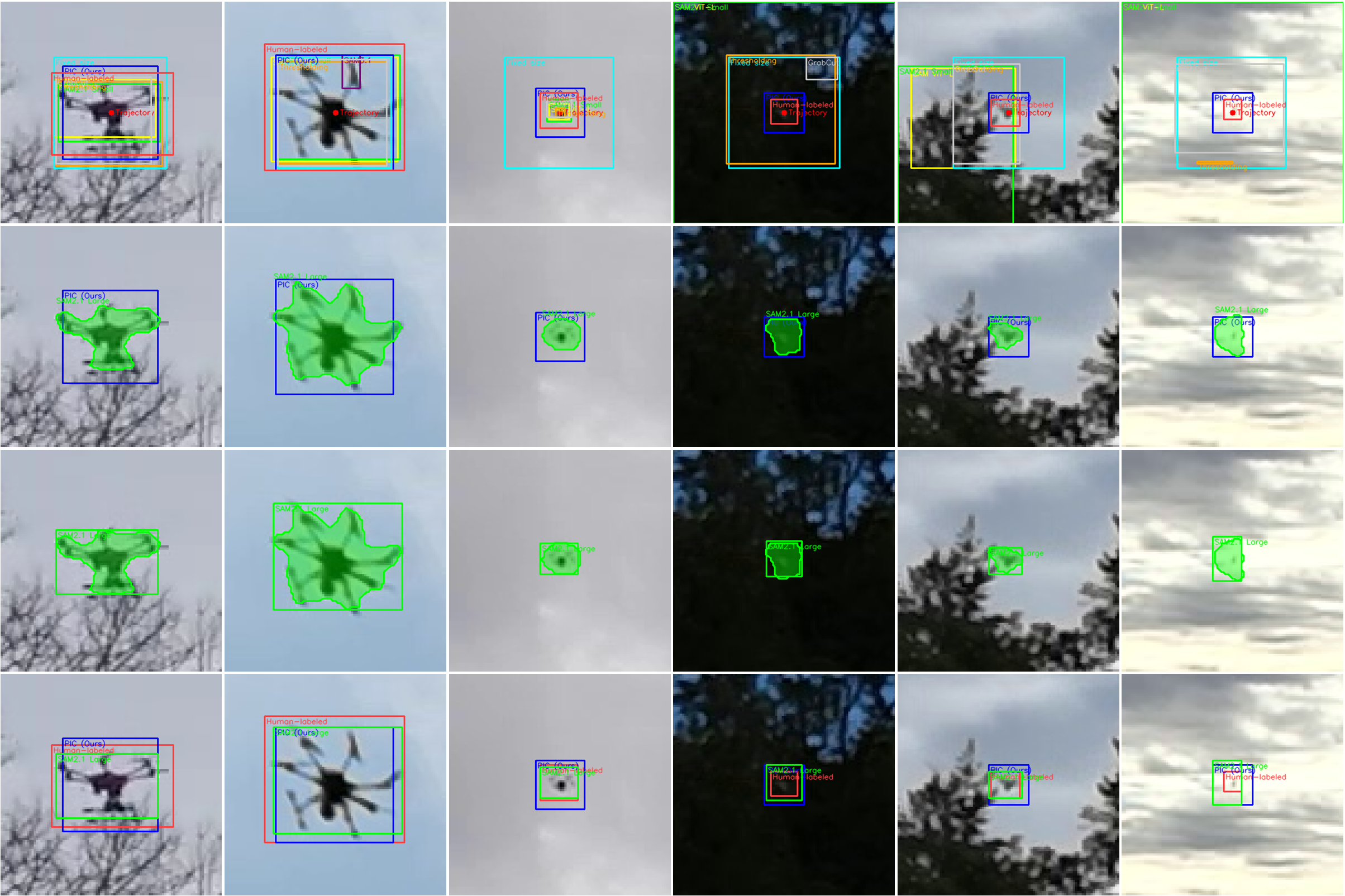}
    \end{center}
    \caption{
    Comparison of bounding boxes generated by different methods.
    \textbf{Top row:} Bounding boxes obtained from human-labeled, fixed-size, image thresholding~\cite{al2010image}, GrabCut~\cite{rother2004grabcut}, SAM-ViT-L~\cite{kirillov2023segment}, SAM2.1-Small~\cite{ravi2024sam}, SAM3.1~\cite{carion2025sam}, and the proposed PIC, with the input point prompt indicated by a red dot.
    \textbf{Second row:} Segmentation masks from SAM2.1 Large~\cite{ravi2024sam} using PIC-generated box prompt.
    \textbf{Third row:} Minimum enclosing bounding boxes derived from the second-row masks.
    \textbf{Bottom row:} Comparison of bounding boxes from human-labeled, PIC, and PIC-SAM2.1-Large.
    }
    \label{fig:overview}
\end{figure*}

Generating high-quality bounding box annotations for UAVs of varying sizes in video data using only trajectory information is a critical step. While learning-based approaches may yield accurate results, they typically require careful model design and training effort. In contrast, we focus on simple, out-of-the-box strategies for bounding box extraction to reduce complexity.
A straightforward solution is to assign fixed-size boxes centered at trajectory points, but this lacks adaptability to UAV scale variations. A natural alternative is to segment local regions around each point and derive bounding boxes from the resulting masks. Classical methods such as image thresholding~\cite{al2010image} are often used for this purpose, but they are sensitive to low-contrast scenes and require manual parameter tuning. GrabCut~\cite{rother2004grabcut} improves segmentation via iterative refinement, yet remains computationally expensive for large-scale annotation. Deep learning-based variants such as DeepGrabCut~\cite{xu2017deep} further increase computational cost.
More recent foundation models such as SAM~\cite{kirillov2023segment,ravi2024sam,carion2025sam} enable zero-shot segmentation with point prompts; however, their performance degrades in UAV scenarios due to domain shift and the inherent spatial noise in trajectory points, often resulting in unstable or inaccurate masks. These limitations are illustrated in the top row of Fig.~\ref{fig:overview}, which compares bounding boxes generated by different methods against human-labeled annotations across diverse datasets and viewpoints.

%
%
%
\section{Methodology}
\label{sec:method}
To construct UAVDB with minimal manual effort, we propose an annotation pipeline that transforms 2D trajectory points into high-quality masks. The pipeline consists of two stages: (1) bounding box generation via PIC, and (2) mask generation using SAM2~\cite{ravi2024sam}.

\subsection{Bounding Box Generation via PIC}
\label{ssec:pic}
The PIC technique extracts UAV bounding boxes from trajectory annotations via an adaptive inward–outward expansion strategy, enabling efficient localization without relying on external models or predefined dimensions. It consists of four steps: (1) initialization, (2) iterative expansion, (3) patch intensity computation, and (4) convergence assessment.

\subsubsection{Initialization}
\label{sssec:initialization}
Given a trajectory point \( (x_0, y_0) \), the bounding box is initialized as a square region \( B_0 \) of size \( w_0 \times h_0 \):
\[
B_0 = \{ (x, y) \mid x_0 - w_0/2 \leq x \leq x_0 + w_0/2, \: y_0 - h_0/2 \leq y \leq y_0 + h_0/2 \}.
\]

\subsubsection{Iterative Expansion}
\label{sssec:expansion}
At each step \(t\), the bounding box expands outward by a fixed size \( \delta \) in all directions:
\[
w_{t+1} = w_t + \delta, \quad h_{t+1} = h_t + \delta, \quad t=0, 1, \ldots
\]  
The expanded region \( B_{t+1} \) captures a progressively larger area around the trajectory point.

\subsubsection{Patch Intensity Computation}
\label{sssec:calculation}
The mean pixel intensity at each step inside the bounding box is computed as:
\[
\mu_t = \frac{1}{|B_t|} \sum_{(x,y) \in B_t} I(x, y).
\]
where \(I(x,y)\) denotes the pixel intensity at \( (x, y) \).

\subsubsection{Convergence Assessment}
\label{sssec:criterion}
Expansion halts when the intensity change between consecutive iterations falls below a threshold \( \epsilon \):  
\[
|\mu_{t+1} - \mu_t| < \epsilon.
\]  
This criterion indicates that further expansion no longer contributes to capturing UAV-relevant pixels, thereby defining the final bounding box boundary.

\begin{figure}[h]
    \begin{center}
    \includegraphics[width=0.85\linewidth]{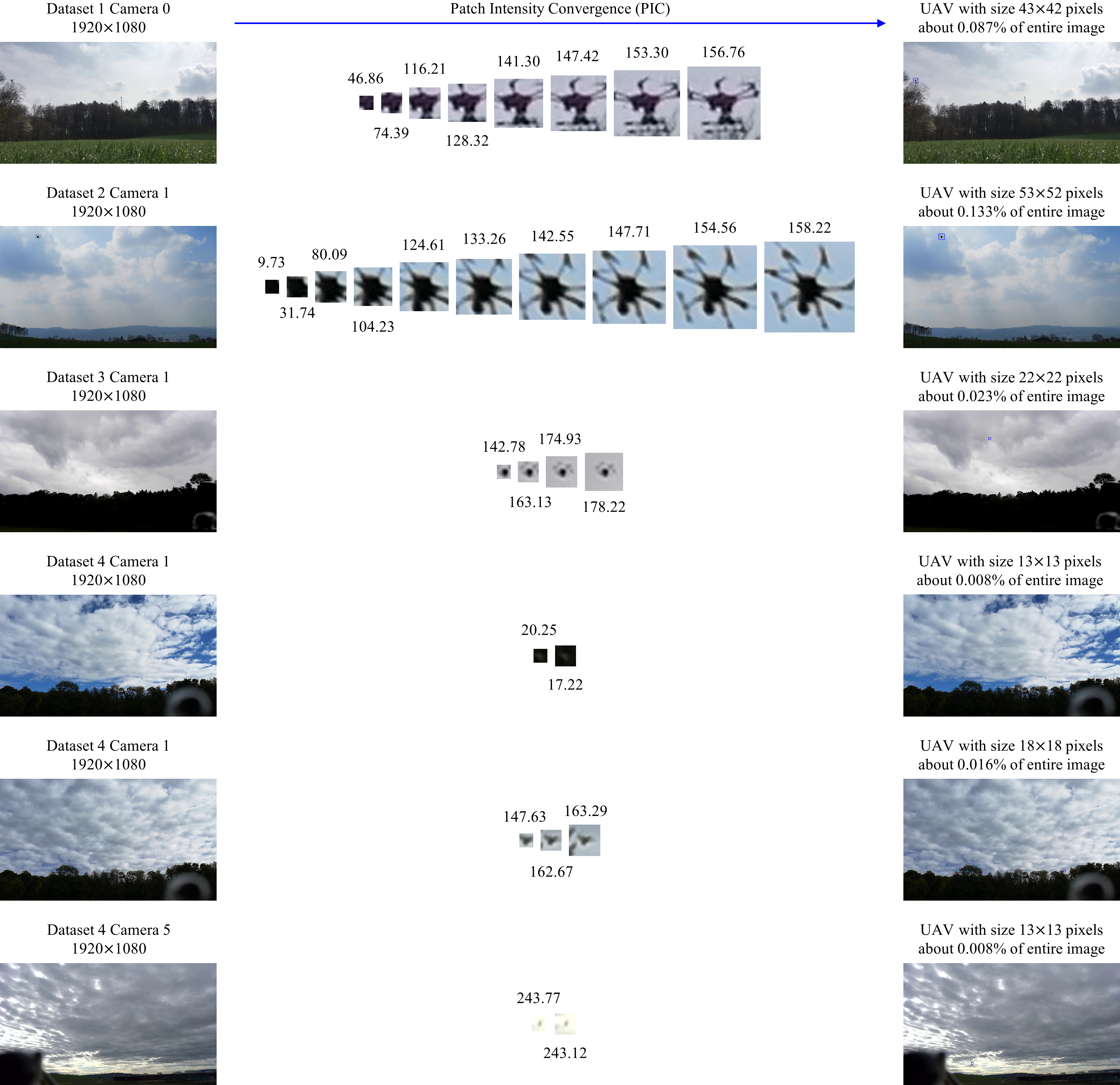}
    \end{center}
    \caption{
    Step-by-step illustration of the PIC process across datasets and camera views. The middle panels show iterative bounding box expansion with corresponding intensity values. The rightmost column presents the final PIC annotations, including UAV size and aspect ratio for each scenario.
    }
    \label{fig:PIC}
\end{figure}

We apply the PIC method to videos and trajectory data from~\cite{li2020reconstruction}, using an initial patch size of \( w_0 = h_0 = 8 \) pixels, an expansion step of \( \delta = 5 \) pixels, and a convergence threshold of \( \epsilon = 4 \). As shown in Fig.~\ref{fig:PIC}, the middle panels visualize the stepwise expansion process and corresponding pixel intensity values across different datasets, demonstrating the robustness of PIC under challenging conditions. The rightmost column provides reference images indicating UAV size as a percentage of the total image area.
PIC successfully localizes UAVs across a wide range of scales, from large instances (53$\times$52 pixels, approximately 0.133\% of the image) to extremely small ones (13$\times$13 pixels, approximately 0.008\%), resulting in high-fidelity bounding box annotations.
For UAVDB, we sample one frame every ten frames (approximately 10\% of the footage) from the sequences listed in Tab.~\ref{tab:dataset}, resulting in 10,763 training images, 2,720 validation images, and 4,578 test images, as summarized in Tab.~\ref{tab:uavdb}. Dataset 5 from~\cite{li2020reconstruction}, which lacks 2D trajectory annotations, is treated as an unseen scenario, with segmentation results presented in the experimental section. Notably, our framework allows flexible control of the frame sampling rate, enabling the dataset scale to be adapted to different application requirements.

\begin{table}[tb]
\caption{Overview of the UAVDB constructed using the proposed PIC approach. The table shows the distribution of images across different datasets and camera configurations, specifying the number of images used for training, validation, and testing.}
\label{tab:uavdb}
\begin{center}
\setlength{\tabcolsep}{8pt}
\resizebox{\linewidth}{!}{$
\begin{tabular}{ccccc}
\toprule
Camera $\backslash$ Dataset & 1 & 2 & 3 & 4 \\
\midrule
0 & train / 291\:\: & test\:\: / 237\:\: & train / 3190 & test\:\: / 2355 \\
1 & valid / 303\:\: & train / 343\:\: & train / 841\:\: & train / 416\:\: \\
2 & train / 394\:\: & train / 809\:\: & valid / 1067 & train / 701\:\: \\
3 & test\:\: / 348\:\: & valid / 426\:\: & train / 638\:\: & train / 727\:\: \\
4 & -- & -- & test\:\: / 1253 & valid / 924\:\: \\
5 & -- & -- & train / 1303 & train / 1110 \\
6 & -- & -- & -- & test\:\: / 385\:\: \\
\bottomrule
\end{tabular}
$}
\end{center}
\end{table}

\subsection{Mask Generation using SAM2}
\label{ssec:sam2}
To extend UAVDB with segmentation annotations, we leverage SAM2~\cite{ravi2024sam}, a powerful zero-shot segmentation model capable of generating instance masks from point or box prompts, inspired in part by~\cite{mukherjee2025common}. In our framework, PIC-generated bounding boxes are used as box prompts to guide SAM2, enabling mask extraction across diverse scenes.
Box-based prompting is essential in our setting. Although SAM2 supports point prompts, we observe that trajectory-derived points are often spatially noisy due to motion blur, occlusion, and annotation inaccuracies. Moreover, without spatial constraints, point prompts frequently lead to unstable or inaccurate masks, particularly for small UAVs, as shown in the top row of Fig.~\ref{fig:overview}.
In contrast, PIC-derived boxes provide spatially localized priors that constrain the segmentation region, allowing SAM2 to focus on relevant areas and produce more accurate masks. As shown in the second and third rows of Fig.~\ref{fig:overview}, PIC-SAM2 consistently yields masks and derived boxes that better capture object shape than PIC bounding boxes, especially for larger UAVs.
However, as illustrated in the rightmost example of the third row in Fig.~\ref{fig:overview}, mask quality may degrade for extremely small UAVs. Nevertheless, PIC-SAM2 does not underperform compared to PIC-based bounding boxes in these cases. As reported in Tab.~\ref{tab:iou_runtime}, PIC-SAM2.1-Large achieves the highest mIoU and is adopted as our mask generation setting.
%
%
%
\section{Experimental Results}
\label{sec:results}
We first compare mIoU and runtime efficiency. We then present comprehensive benchmark results on UAVDB using YOLO-series detectors.

\subsection{Annotation Accuracy and Runtime Efficiency}
\label{ssec:annotation}
We manually annotate a subset of UAVDB and use the resulting human-labeled annotations as ground-truth. For the fixed-size and thresholding~\cite{al2010image} approaches, we adopt a 50$\times$50 region and set the threshold to 150, based on empirical tuning for optimal performance. GrabCut~\cite{rother2004grabcut}, SAM~\cite{kirillov2023segment}, and SAM2~\cite{ravi2024sam} are implemented using OpenCV and their official pretrained models, respectively.
Despite the advanced segmentation capabilities of SAM and SAM2, direct point prompting yields a low mIoU due to the presence of unconstrained regions.
As reported in the left part of Tab.~\ref{tab:iou_runtime}, the proposed PIC method achieves the highest mIoU among point-prompt-based approaches, with a runtime of only 0.007 seconds, introducing negligible computational overhead beyond image I/O. In contrast, manual annotation requires an average of 19 seconds per bounding box, making it impractical for large-scale datasets with small objects.

\begin{table}[tb]
\caption{Comparison of bounding box generation in terms of mIoU and runtime.}
\label{tab:iou_runtime}
\centering
\setlength{\tabcolsep}{8pt} 
\begin{tabular}{lcccc}
\toprule
\multirow{2.5}{*}{Method} & \multicolumn{2}{c}{Point prompt} & \multicolumn{2}{c}{PIC box prompt} \\
\cmidrule{2-5}
& mIoU \(\uparrow\) & Runtime (s) \(\downarrow\) & mIoU \(\uparrow\) & Runtime (s) \(\downarrow\) \\
\midrule
Human-labeled & 1.000 & 19.00 & -- & -- \\
Fixed-size    & 0.278 & \textbf{0.002}  & -- & -- \\
Thresholding~\cite{al2010image}  & 0.316 & 0.012  & -- & -- \\
GrabCut~\cite{rother2004grabcut} & 0.425 & 2.192  & -- & -- \\
PIC (Ours)    & \textbf{0.462} & 0.007  & -- & -- \\
\midrule
SAM~\cite{kirillov2023segment} & & & & \\
\quad ViT-B & 0.233 & 0.505 & 0.451 & 0.497 \\
\quad ViT-L & 0.238 & 1.162 & 0.436 & 1.171 \\
\quad ViT-H & 0.238 & 4.236 & 0.419 & 4.251 \\
\addlinespace
SAM2.1~\cite{ravi2024sam} & & & & \\
\quad Tiny & 0.239 & 0.092 & 0.496 & \textbf{0.079} \\
\quad Small & 0.254 & 0.077 & 0.478 & 0.084 \\
\quad Base+ & 0.251 & 0.121 & 0.476 & 0.128 \\
\quad Large & 0.244 & 0.234 & \textbf{0.582} & 0.238 \\
\addlinespace
SAM3 Models~\cite{carion2025sam} & & & & \\
\quad SAM3 & -- & -- & 0.284 & 0.718 \\
\quad SAM3.1 & -- & -- & 0.368 & 0.573 \\
\bottomrule
\end{tabular}
\end{table}

\subsection{Benchmark on UAVDB}
\label{ssec:benchmark}

We benchmark UAVDB using YOLO-series models ranging from YOLOv8 to YOLO26~\cite{Jocher_YOLO_by_Ultralytics_2023,wang2024yolov9,wang2024yolov10,yolo11_ultralytics,tian2025yolov12attentioncentricrealtimeobject,lei2025yolov13,yolo26_ultralytics}. All experiments are conducted on an HPC system~\cite{meade2017spartan} equipped with an NVIDIA A100 GPU (80\,GB memory).
For object detection, models are trained with an input size of 640 and a batch size of 32. For instance segmentation, models are trained with an input size of 1280 and a batch size of 16. Each model is fine-tuned from its official pretrained weights for 100 epochs using eight dataloader workers. Mosaic augmentation is applied throughout training, except for the final 10 epochs.
Tab.~\ref{tab:results} reports training time, inference speed, model size, and AP on the validation and test sets.

\begin{figure*}[tb]
    \begin{center}
    \includegraphics[width=\linewidth]{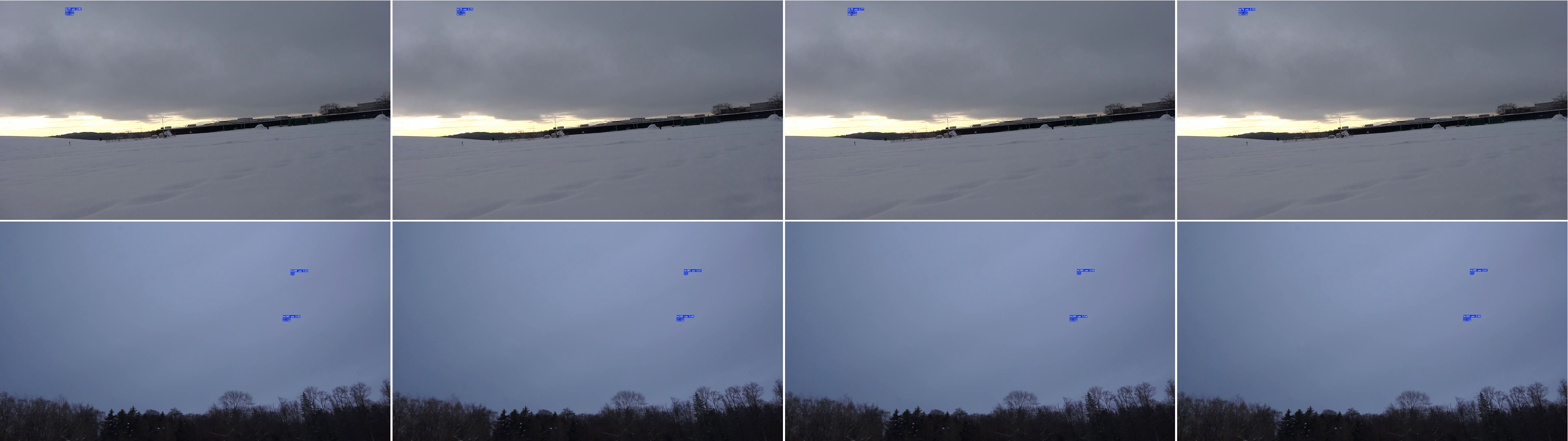}
    \end{center}
    \caption{Sequential tracking results from YOLOv12n-seg~\cite{tian2025yolov12attentioncentricrealtimeobject} on unseen Dataset 5. Top: Camera 3; Bottom: Camera 5. Frames shown left to right.}
    \label{fig:predict_d5}
\end{figure*}

We further visualize the generalization capability of a trained YOLOv12n-seg model on Dataset 5, which is entirely excluded from training and validation. Unlike typical unseen splits with similar data distributions, Dataset 5 represents a distinct scenario, making detection and segmentation more challenging.
As shown in Fig.~\ref{fig:predict_d5}, we present sequential predictions from Camera 3 (top row) and Camera 5 (bottom row) across consecutive frames. Despite the UAVs being small, blurry, and often embedded in complex backgrounds, the model demonstrates strong generalization, producing well-aligned bounding boxes and tightly fitted segmentation masks.
Leveraging the video-based nature of UAVDB, we extend beyond static detection to continuous tracking, enabling more comprehensive and realistic evaluation compared to image-level detection.

\begin{table*}[tb]
\caption{Performance comparison of YOLOv8 to YOLO26~\cite{Jocher_YOLO_by_Ultralytics_2023,wang2024yolov9,wang2024yolov10,yolo11_ultralytics,tian2025yolov12attentioncentricrealtimeobject,lei2025yolov13,yolo26_ultralytics} on UAVDB using PIC-generated bounding boxes for object detection and PIC-SAM2.1-Large masks for instance segmentation. Bold indicates the best performance, while underline denotes the most practical model in terms of size or inference speed.}
\label{tab:results}
\small
\begin{center}
\begin{adjustbox}{max width=\linewidth}
\begin{tabular}{lccrrcccc}
\toprule
\multirow{2}{*}{Model} & Training Time & Inference Time & \multirow{2}{*}{\#Param. (M)} & \multirow{2}{*}{FLOPs (G)} & \multirow{2}{*}{$\text{AP}^{val}_{50}$} & \multirow{2}{*}{$\text{AP}^{val}_{50-95}$} & \multirow{2}{*}{$\text{AP}^{test}_{50}$} & \multirow{2}{*}{$\text{AP}^{test}_{50-95}$}\\
& (hours:mins:sec) & (per image, ms) & & & \\
\midrule
\multicolumn{9}{c}{\textit{Object Detection}} \\
\midrule
YOLOv8n & \underline{01:40:31} & 0.9 & 2.685\hspace*{4mm} & 6.8\hspace*{4mm} & 0.829 & 0.522\hspace*{4mm} & 0.789\hspace*{2mm} & 0.450 \\
YOLOv8s & 01:55:05 & 1.2 & 9.828\hspace*{4mm} & 23.3\hspace*{4mm} & 0.814 & 0.545\hspace*{4mm} & 0.796\hspace*{2mm} & 0.450 \\
YOLOv8m & 02:43:08 & 1.8 & 23.203\hspace*{4mm} & 67.4\hspace*{4mm} & 0.809 & 0.538\hspace*{4mm} & 0.827\hspace*{2mm} & 0.526 \\
YOLOv8l & 03:54:44 & 2.6 & 39.434\hspace*{4mm} & 145.2\hspace*{4mm} & 0.830 & 0.563\hspace*{4mm} & 0.836\hspace*{2mm} & 0.544 \\
YOLOv8x & 04:33:08 & 3.5 & 61.597\hspace*{4mm} & 226.7\hspace*{4mm} & 0.820 & 0.554\hspace*{4mm} & 0.728\hspace*{2mm} & 0.448 \\
\addlinespace
YOLOv9t & 02:53:11 & 2.5 & 2.617\hspace*{4mm} & 10.7\hspace*{4mm} & 0.839 & 0.501\hspace*{4mm} & 0.848\hspace*{2mm} & 0.508 \\
YOLOv9s & 03:05:02 & 2.6 & 9.598\hspace*{4mm} & 38.7\hspace*{4mm} & 0.819 & 0.517\hspace*{4mm} & 0.834\hspace*{2mm} & 0.484 \\
YOLOv9m & 05:08:28 & 4.1 & 32.553\hspace*{4mm} & 130.7\hspace*{4mm} & 0.840 & 0.507\hspace*{4mm} & 0.858\hspace*{2mm} & 0.522 \\
YOLOv9c & 06:17:08 & 5.3 & 50.698\hspace*{4mm} & 236.6\hspace*{4mm} & 0.851 & 0.544\hspace*{4mm} & 0.851\hspace*{2mm} & 0.504 \\
YOLOv9e & 08:00:05 & 6.6 & 68.548\hspace*{4mm} & 240.7\hspace*{4mm} & 0.755 & 0.414\hspace*{4mm} & 0.768\hspace*{2mm} & 0.383 \\
\addlinespace
YOLOv10n & 02:05:39 & \underline{0.7} & 2.695\hspace*{4mm} & 8.2\hspace*{4mm} & 0.764 & 0.492\hspace*{4mm} & 0.731\hspace*{2mm} & 0.417 \\
YOLOv10s & 02:23:03 & 1.2 & 8.036\hspace*{4mm} & 24.4\hspace*{4mm} & 0.817 & 0.530\hspace*{4mm} & 0.823\hspace*{2mm} & 0.516 \\
YOLOv10m & 03:06:59 & 1.8 & 16.452\hspace*{4mm} & 63.4\hspace*{4mm} & 0.798 & 0.531\hspace*{4mm} & 0.821\hspace*{2mm} & 0.536 \\
YOLOv10b & 03:29:18 & 2.1 & 20.413\hspace*{4mm} & 97.9\hspace*{4mm} & 0.801 & 0.517\hspace*{4mm} & 0.760\hspace*{2mm} & 0.467 \\
YOLOv10l & 04:04:22 & 2.5 & 25.718\hspace*{4mm} & 126.3\hspace*{4mm} & 0.774 & 0.502\hspace*{4mm} & 0.842\hspace*{2mm} & 0.517 \\
YOLOv10x & 05:14:07 & 3.5 & 31.586\hspace*{4mm} & 169.8\hspace*{4mm} & 0.771 & 0.507\hspace*{4mm} & 0.693\hspace*{2mm} & 0.431 \\
\addlinespace
YOLO11n & 01:50:00 & 0.9 & 2.582\hspace*{4mm} & 6.3\hspace*{4mm} & 0.847 & 0.527\hspace*{4mm} & \underline{0.856}\hspace*{2mm} & \underline{0.539} \\
YOLO11s & 02:07:01 & 1.2 & 9.413\hspace*{4mm} & 21.3\hspace*{4mm} & 0.826 & 0.553\hspace*{4mm} & 0.885\hspace*{2mm} & 0.578 \\
YOLO11m & 03:07:40 & 1.9 & 20.031\hspace*{4mm} & 67.6\hspace*{4mm} & 0.827 & 0.588\hspace*{4mm} & 0.843\hspace*{2mm} & 0.578 \\
YOLO11l & 04:09:45 & 2.4 & 25.280\hspace*{4mm} & 86.6\hspace*{4mm} & 0.810 & 0.555\hspace*{4mm} & 0.798\hspace*{2mm} & 0.517 \\
YOLO11x & 05:20:38 & 3.6 & 56.828\hspace*{4mm} & 194.4\hspace*{4mm} & 0.812 & 0.560\hspace*{4mm} & 0.782\hspace*{2mm} & 0.534 \\
\addlinespace
YOLOv12n & 02:15:38 & 1.8 & 2.557\hspace*{4mm} & 6.3\hspace*{4mm} & \underline{0.857} & 0.544\hspace*{4mm} & 0.848\hspace*{2mm} & 0.531 \\
YOLOv12s & 02:44:29 & 2.0 & 9.231\hspace*{4mm} & 21.2\hspace*{4mm} & 0.869 & 0.566\hspace*{4mm} & 0.882\hspace*{2mm} & 0.565 \\
YOLOv12m & 03:34:36 & 2.6 & 20.106\hspace*{4mm} & 67.1\hspace*{4mm} & 0.866 & 0.567\hspace*{4mm} & 0.886\hspace*{2mm} & 0.584 \\
YOLOv12l & 05:10:15 & 3.1 & 26.340\hspace*{4mm} & 88.5\hspace*{4mm} & 0.870 & 0.584\hspace*{4mm} & 0.875\hspace*{2mm} & 0.590 \\
YOLOv12x & 06:35:47 & 3.9 & 59.045\hspace*{4mm} & 198.5\hspace*{4mm} & 0.879 & 0.576\hspace*{4mm} & \textbf{0.896}\hspace*{2mm} & 0.569 \\
\addlinespace
YOLOv13n & 03:23:00 & 1.6 & 2.448\hspace*{4mm} & 6.2\hspace*{4mm} & 0.833 & 0.541\hspace*{4mm} & 0.795\hspace*{2mm} & 0.505 \\
YOLOv13s & 04:15:04 & 2.1 & 9.530\hspace*{4mm} & 21.3\hspace*{4mm} & 0.852 & 0.555\hspace*{4mm} & 0.804\hspace*{2mm} & 0.496 \\
YOLOv13l & 10:07:28 & 5.5 & 27.514\hspace*{4mm} & 88.1\hspace*{4mm} & 0.860 & 0.554\hspace*{4mm} & 0.826\hspace*{2mm} & 0.540 \\
YOLOv13x & 13:40:58 & 8.3 & 63.886\hspace*{4mm} & 198.7\hspace*{4mm} & 0.846 & 0.568\hspace*{4mm} & 0.836\hspace*{2mm} & 0.556 \\
\addlinespace
YOLO26n & 03:13:23 & 0.8 & \underline{2.375}\hspace*{4mm} & \underline{5.2}\hspace*{4mm} & 0.845 & \underline{0.564}\hspace*{4mm} & 0.799\hspace*{2mm} & 0.528 \\
YOLO26s & 03:33:57 & 1.1 & 9.466\hspace*{4mm} & 20.5\hspace*{4mm} & 0.863 & 0.555\hspace*{4mm} & 0.843\hspace*{2mm} & 0.567 \\
YOLO26m & 04:34:29 & 1.8 & 20.350\hspace*{4mm} & 67.8\hspace*{4mm} & 0.834 & 0.577\hspace*{4mm} & 0.776\hspace*{2mm} & 0.533 \\
YOLO26l & 05:53:52 & 2.2 & 24.747\hspace*{4mm} & 86.1\hspace*{4mm} & 0.868 & 0.585\hspace*{4mm} & 0.875\hspace*{2mm} & \textbf{0.594} \\
YOLO26x & 07:18:58 & 3.5 & 55.635\hspace*{4mm} & 193.4\hspace*{4mm} & \textbf{0.890} & \textbf{0.595}\hspace*{4mm} & 0.775\hspace*{2mm} & 0.530 \\
\midrule
\multicolumn{9}{c}{\textit{Instance Segmentation}} \\
\midrule
\multirow{2}{*}{YOLOv12n-seg} & \multirow{2}{*}{10:14:58} & \multirow{2}{*}{4.7} & \multirow{2}{*}{2.761\hspace*{4mm}} & \multirow{2}{*}{9.7\hspace*{4mm}} & \textsuperscript{Box}0.925\hspace*{4mm}  & 0.591\hspace*{4mm} & 0.928\hspace*{2mm} & 0.531 \\
             &          &     &                    &                  & \textsuperscript{Mask}0.860\hspace*{5.5mm} & 0.396\hspace*{4mm} & 0.569\hspace*{2mm} & 0.209 \\
\addlinespace
\multirow{2}{*}{YOLO26n-seg}  & \multirow{2}{*}{09:36:28} & \multirow{2}{*}{3.0} & \multirow{2}{*}{2.689\hspace*{4mm}} & \multirow{2}{*}{9.0\hspace*{4mm}} & \textsuperscript{Box}0.922\hspace*{4mm}  & 0.626\hspace*{4mm} & 0.911\hspace*{2mm} & 0.525 \\
             &          &     &                    &                  & \textsuperscript{Mask}0.847\hspace*{5.5mm} & 0.403\hspace*{4mm} & 0.562\hspace*{2mm} & 0.208 \\
\bottomrule
\end{tabular}
\end{adjustbox}
\end{center}
\end{table*}

\clearpage

\subsection{Discussion}
\label{ssec:analysis_and_discussion}

First, we discuss the sensitivity of PIC to its three parameters: the initial patch size \((w_0, h_0)\), expansion step \(\delta\), and convergence threshold \(\epsilon\). A larger initial patch accelerates convergence but may include background clutter, whereas a smaller patch improves localization of tiny UAVs at the cost of more iterations. Similarly, a larger step size \(\delta\) reduces computation but may overshoot object boundaries, while a smaller step size yields tighter localization at higher computational cost. The convergence threshold \(\epsilon\) controls the stopping criterion: stricter thresholds improve bounding-box fidelity but yield diminishing returns in accuracy. Overall, these trade-offs indicate that PIC can be effectively tuned by aligning parameter scales with object size and motion characteristics across datasets.

Another limitation of PIC is that it tends to produce near-square bounding boxes, which may not align well with elongated UAVs or objects in other domains. This limitation can be mitigated by the SAM2 refinement stage. As shown in the bottom row of Fig.~\ref{fig:overview}, PIC-SAM2.1-Large outputs are no longer constrained to square shapes. This demonstrates that the segmentation stage corrects systematic misalignment by converting coarse square proposals into more accurate object shapes, thereby improving annotation quality.
In addition, as shown in the segmentation results in Tab.~\ref{tab:results}, the performance gap between the validation and test sets suggests potential overfitting. This issue can be alleviated by increasing the dataset size, which is straightforward in UAVDB due to its flexible frame extraction mechanism.
%
%
%
\section{Conclusion}
\label{sec:conclusion}
We introduced UAVDB, a UAV benchmark designed for RGB-based camera-to-UAV monitoring in long-range aerospace surveillance scenarios. UAVDB is built upon a PIC-SAM2 pipeline that generates high-quality instance masks with minimal human effort. Beyond detection and segmentation, its video-based nature supports flexible scaling via adjustable frame sampling and enables temporal tasks such as tracking, making it more versatile than static image benchmarks.
Furthermore, the pipeline is transferable and can be readily integrated into other point-guided vision tasks. We expect the proposed annotation pipeline to further benefit research in weakly supervised computer vision.
%
%
%
\subsubsection{Acknowledgements}
We thank the University of Melbourne and National Yang Ming Chiao Tung University for providing computational resources. We also gratefully acknowledge all open-source projects and tools used in this work.

%
%
%
\bibliographystyle{splncs04}
\bibliography{mybibliography}
%




\end{document}